\documentclass[jkps,preprint,fleqn,showpacs,showkeys]{revtex4}
\usepackage{graphicx}
\usepackage{amssymb}
\usepackage{amsmath}
\usepackage{bm}
\begin{document}
\setcounter{page}{0}
\title[]{Comparisons of Wavelet Functions in QRS Signal to Noise Ratio Enhancement and Detection Accuracy}
\author{Pornchai \surname{PHUKPATTARANONT}}
\email{pornchai.p@psu.ac.th}
\thanks{Tel: +66-7455-8831 Fax: +66-7445-9395}
\affiliation{Department of Electrical Engineering, Prince of Songkla University, Songkhla 90110, Thailand.}

\date[]{Received 16 April 2015}

\begin{abstract}
We compare the capability of wavelet functions used for noise removal in preprocessing step of a QRS detection algorithm in the electrocardiogram (ECG) signal. The QRS signal to noise ratio enhancement and the detection accuracy of each wavelet function are evaluated using three measures: (1) the ratio of the maximum beat amplitude to the minimum beat amplitude (RMM), (2) the mean of absolute of time error (MATE), and (3) the figure of merit (FOM). Three wavelet functions from previous well-known publications are explored, i.e., Bior1.3, Db10, and Mexican hat wavelet functions. Results evaluated with the ECG signal from MIT-BIH arrhythmia database show that the Mexican hat wavelet function is better than the others. While the scale 8 of Mexican hat wavelet function can provide the best enhancement in QRS signal to noise ratio, the scale 4 of Mexican hat wavelet function can provide the best detection accuracy. These results may be combined and may enable the use of a single fixed threshold for all ECG records leading to the reduction in computational complexity of the QRS detection algorithm.     
\end{abstract}

\pacs{05.45.Tp, 05.45Df}

\keywords{Electrocardiography (ECG), Signal to noise ratio,  QRS detection, Signal processing, Continuous wavelet transform}

\maketitle

\section{Introduction} 
The detection of the heart disease is very important. One of the measurements used for checking the condition of the heart is the electrocardiogram (ECG) signal. The components of each ECG beat consist of the P wave, the QSR complex, and the S wave. However, the manual analysis of ECG signal is tedious and time consuming. Therefore, a computer-assisted system for analyzing ECG signals has been developed to solve these problems. The detection algorithm for R peak in the QRS complex is a preliminary step in the computer-assisted system to other ECG signal analysis such as arrhythmia analysis \cite{1,2}, and heart rate variability (HRV) analysis \cite{3}.    

The algorithm for detecting R peak in the QRS complex is generally divided into three main steps, i.e. noise removal, envelope signal determination, and R peak detection. In noise removal step, a variety of noises in the ECG signal, such as the muscle noise, the power line noise, and the baseline drift noise are removed. one of the most popular methods used for noise removal in the ECG signal is the wavelet transform technique including continuous wavelet transform (CWT) \cite{4,5,6,7}, discrete wavelet transform \cite{8,9,10,11,12,13,14}, and wavelet package decomposition \cite{15}. After noise removal, the envelope signal is obtained from the ECG signal after noise removal and the threshold is defined to determine the time period where the R peak in QRS complex locates. Generally, there are four types of thresholding techniques, i.e., a single level fixed threshold \cite{7}, multiple levels fixed threshold \cite{10,11,12,15}, a single level adaptive threshold \cite{4,5,6,8,9,13,14}, and multiple levels adaptive threshold. The R peak can be detected from the time position where the amplitude of the ECG signal after noise removal is maximal.          

Generally, more complicated threshold techniques such as multiple levels fixed threshold and the single level adaptive threshold are used because the ECG signals after noise removal from different beat types, such as normal beat and  premature ventricular contraction beat, have significant difference in their amplitudes due to their difference in frequency components. All wavelet functions in previous publications were selected based on their efficiency in removing noise. However, their effects on QRS signal to noise ratio and the detection accuracy are not yet considered. Therefore, the proposed comparison measurements of the wavelet function in CWT are presented in this paper to demonstrate both QRS signal to noise ratio enhancement and detection accuracy so that the reduction in the computational complexity of the algorithm in the R peak detection step can be obtained.     

The rest of this paper is organized as follows. Section 2 presents the CWT theory. Section 3 describes the QRS detection algorithm, the performance measures for the selection of wavelet functions, as well as the measures for detection accuracy. Results and discussion are given in Section 4. Finally, conclusions are drawn in Section 5.

\section{Continuous Wavelet Transform} 
The CWT has been gained popular uses for decomposing signals in many applications including noise removal in ECG signals. Given the input signal $x(t)$, which is the ECG signal in this paper, the CWT of $x(t)$ can be expressed as

\begin{equation}\label{CWT}
T_{a,b} =  \int_{-\infty}^{\infty}x(t)\frac{1}{\sqrt{a}}\psi^{*}\left( \frac{t-b}{a} \right)dt,
\end{equation}

where $T_{a,b}$ is the CWT of $x(t)$, $a$ is the dilation or scale parameter, $b$ is the location parameter, and $\psi^{*}(t)$ is the complex conjugate of the wavelet function. The scale and the wavelet function are two important parameters affecting the performance of noise removal in ECG signals. There are various types of the wavelet functions that are successfully used for removing noises in ECG signals from previous publications including Bior1.3 \cite{5}, Db10 \cite{14}, and Mexican hat wavelet \cite{6} functions. This paper proposes to study and quantitatively compare the performance of these three wavelet functions on their capability in enhancing QRS signal to noise ratio and demonstrate its potential application in QRS detection algorithm of ECG data.        

\section{Materials and Methods}
\subsection{QRS Detection Algorithm}
\begin{figure}
\centering
\includegraphics[width=0.7\linewidth]{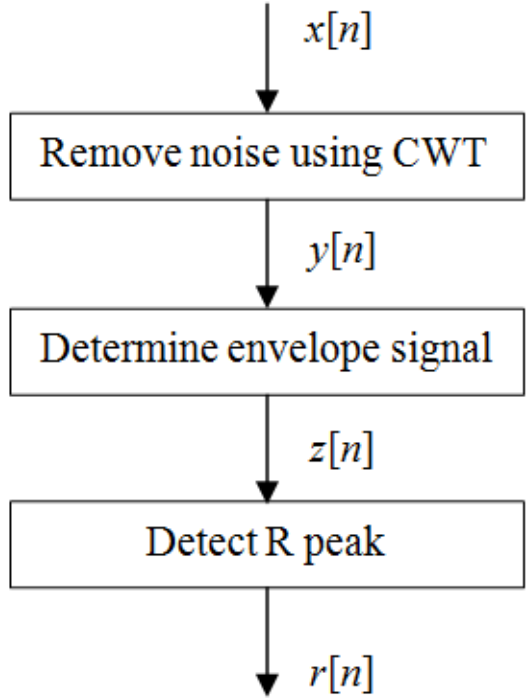}
\caption{Block diagram of the QRS detection algorithm used for evaluating wavelet functions.}
\label{fig:schematic}
\end{figure}
Fig. \ref{fig:schematic} shows a block diagram of the QRS detection algorithm used for evaluating wavelet functions, which consists of three steps: noise removal using CWT, envelope signal determination, and R peak detection. Details of each step are as follows.

\begin{enumerate}
\item) Determine the signal after noise removal $y[n]$ from the ECG signal $x[n]$ by processing based on the CWT, which can be expressed as 
\begin{equation}
y[n] = T^2_{a,b}.
\end{equation}
\item) Determine the envelope signal $z[n]$ from $y[n]$ using the maximum filter with the length $L$ = 120 ms as given by
\begin{equation}\label{svf1}
z[n] = \max_{k\in[n-L+1,n]}{y[k]}.
\end{equation}
 \item) Detect the R peak $r[n]$ in QRS complex using the following steps. 
\begin{enumerate}
\item Define a threshold value $thv$.
\item Find the time duration where $z[n]$ is greater than $thv$ and determine the beginning time $t_1$ and the ending time $t_2$.
\item Determine the R peak location $t_R$ from time between $[t_1 \;\; t_2]$ in $y[n]$ that gives the maximum amplitude.
\end{enumerate}
\end{enumerate}

\begin{figure}
\centering
     \includegraphics[width=0.8\linewidth]{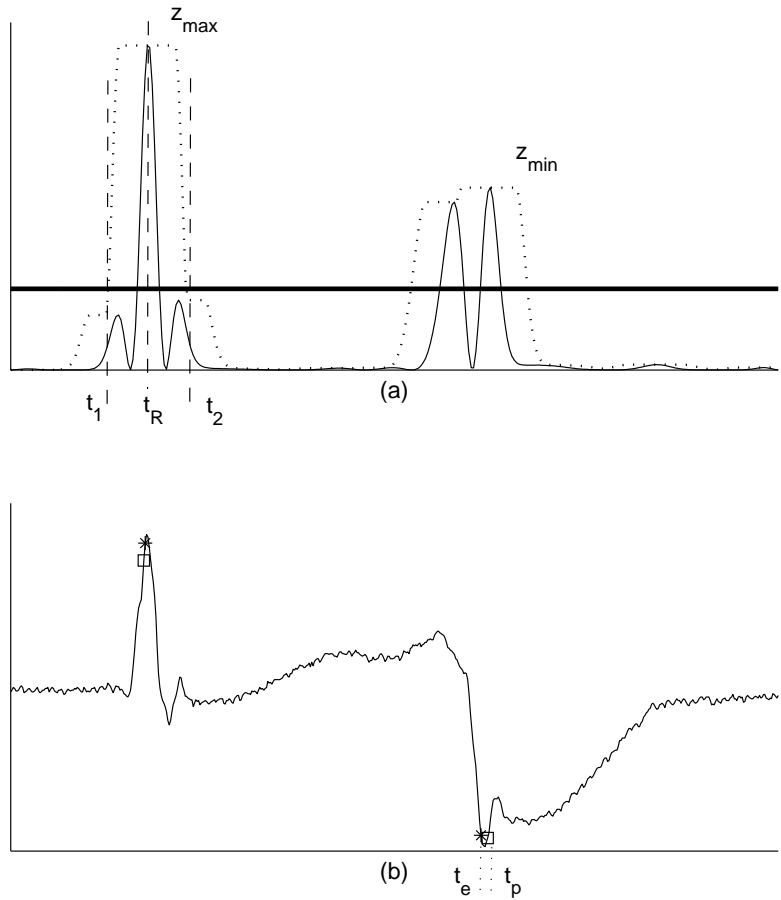}
    \caption{Signal characteristics in QRS detection algorithm. (a) Thin line: ECG signal after noise removal $y[n]$. 
    Dotted line: Envelope signal $z[n]$. Thick line: Threshold value line. (b) ECG signal overlaid by the markers from the algorithm (square) and the expert (asterisk). While the signal on the left hand side is a normal ECG beat, the signal on the right hand side is a premature ventricular contraction beat.} 
    \label{CIN00}
\end{figure}
Fig. \ref{CIN00} shows an example of signal characteristics determined from the QRS detection algorithm. Fig. \ref{CIN00}(b) shows an example of two beats of the ECG signal $x[n]$. While the beat on the left hand side is a normal ECG beat, the beat on the right hand side is a premature ventricular contraction beat. The noises in the ECG signal $x[n]$ are removed based on the CWT to obtain $y[n]$. Fig. \ref{CIN00}(a) shows an example of the ECG signal after noise removal $y[n]$ as described in the first step of the algorithm using a thin line. While the normal beat has a single peak after noise removal, the premature ventricular contraction beat has double peaks. Subsequently, the envelope signal $z[n]$ is calculated from $y[n]$ as described in the second step of the algorithm and is shown in Fig. \ref{CIN00}(a) using a dotted line. Then, the threshold value $thv$ is defined as shown in Fig. \ref{CIN00}(a) using a thick line. Finally, the R peak in the QRS complex $r[n]$ is determined from the time where the maximum peak occurs within the time period defined by $z[n]$ and the threshold value as described in the third step of the algorithm. Fig. \ref{CIN00}(a) shows an example of the beginning time $t_1$, the ending time $t_2$, and the time $t_R$ where the maximum peak locates. The $z_{max}$ and $z_{min}$ are maximum beat amplitude and minimum beat amplitude, respectively. In addition, the ECG signal overlaid by the markers from the proposed algorithm (square) and the expert (asterisk) is shown in Fig. \ref{CIN00}(b). The time $t_p$ and $t_e$ are the position of R peak given by the algorithm and the position of R peak given by the expert, respectively.   

\subsection{Performance Evaluation for Wavelet Functions}
Three parameters were used in evaluating the performance of the wavelet function in the CWT consisting of the ratio of the maximum beat amplitude $z_{max}$ to the minimum beat amplitude $z_{min}$ (RMM), the sum of absolute of time error (MATE), and the figure of merit (FOM). Their details are as follows.
\begin{itemize}                                                                                                   
\item RMM: The RMM can be expressed as                                                                                                                                                      
\begin{equation}\label{RMMeq}
\textrm{RMM} = \frac{z_{max}}{z_{min}}.
\end{equation}
If the value of RMM is closed to 1, the amplitude of all beats in the envelope signal $z[n]$ is almost equal. In other words, the QRS signal to noise ratio is maximized. As a result, the single fixed threshold technique can be used instead of the adaptive threshold technique. This is important because it can reduce the computational complexity of QRS detection algorithm. 

\item MATE: The MATE in the unit of millisecond (ms) is given by
\begin{equation}\label{MATEeq}
\textrm{MATE} = \frac{1}{N}\sum_{k=1}^{N}|t_p(k)-t_e(k)|,
\end{equation}
 where $N$ is the total number of ECG beats under considerations, $t_p(k)$ is the time where the R peak locates determined from the algorithm and $t_e(k)$ is the time where the R peak locates determined from the expert. The best value of MATE is 0, which means that all R peaks in the QRS complex are correctly detected by the algorithm.

\item FOM: The FOM can be defined by  
\begin{equation}\label{FOMeq}
\textrm{FOM} = \frac{1}{\textrm{RMM}\times \textrm{MATE}}.
\end{equation}
FOM is determined using the combination of RMM and MATE. The higher FOM, the more appropriate for the wavelet function in removing noises from the ECG signal and equalizing the amplitude of ECG beats after processing with the CWT.
\end{itemize}

\subsection{Performance Evaluation for Detection Accuracy}
The performance of QRS detection algorithm is evaluated with sensitivity
(SEN), positive predictive rate (PPR), and detection error rate (DER). SEN is given by
\begin{equation}
\textrm{SEN} = \frac{\textrm{TP}}{\textrm{TP}+\textrm{FN}}\times100,
\end{equation}
where true positive (TP) is the number of correct QRS complex predictions. FN is the false
negative prediction. In other words, the algorithm predicts that there is no QRS complex in the
location where there is a real QRS complex. PPR can be expressed as
\begin{equation}
\textrm{PPR} = \frac{\textrm{TP}}{\textrm{TP}+\textrm{FP}}\times100,
\end{equation}
where FP is the false positive prediction. In other words, the algorithm predicts that
there is a QRS complex in the location where there is no QRS complex. DER is used for
evaluating the accuracy of algorithm including both FN and FP values, which can be
given by
\begin{equation}
\textrm{DER} = \frac{\textrm{FN}+\textrm{FP}}{\textrm{TP}+\textrm{FN}}\times100.
\end{equation}

\section{Results and Discussion}
\subsection{Comparisons of Wavelet Functions}
We measure the performance of wavelet functions Bior1.3, Db10, and Mexican hat with the ECG data record 207 containing the mixture of normal beats and premature ventricular contraction beats from MIT-BIH arrhythmia database \cite{16}. The QRS complex is detected using the ECG signal from channel 1 or lead II only. In other words, the ECG signal from channel 1 was represented by $x[n]$.

\begin{figure}
\centering
\includegraphics[width=0.8\linewidth]{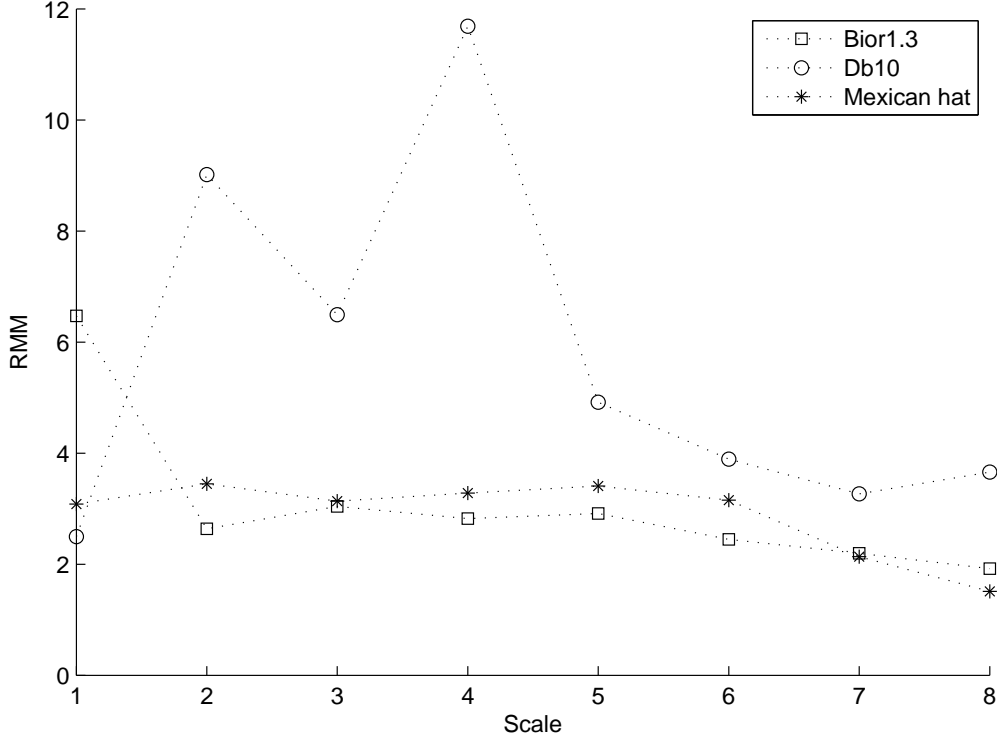}
\caption{Comparisons of the RMM value as a function of scale from three wavelet functions. Results from the Bior1.3, Db10, and Mexican hat wavelet functions are shown using square, circle, and asterisk markers, respectively.}
\label{CIN01}
\end{figure}
Fig. \ref{CIN01} shows the results of RMM as a function of the scale from 1 to 8 applied on the ECG signal record 207 from time 12.6 s to 22.6 s. In other words, the CWT processing method is $T_{a,b}^2$ when the scale $a$ is varied from 1 to 8. Results from the Bior1.3, Db10, and Mexican hat wavelet functions are shown using square, circle, and asterisk markers, respectively. The RMM values from all wavelet functions tend to decrease when the scales increase from 1 to 8. The RMM values from the Db10 wavelet function are greater than those from other wavelet functions at same scale (Except the scale 1). The maximum RMM is 11.69 at the scale 4 of the Db10 wavelet function and the minimum RMM is 1.51 at the scale 8 of the Mexican hat wavelet function.  

\begin{figure}
\centering
\includegraphics[width=0.8\linewidth]{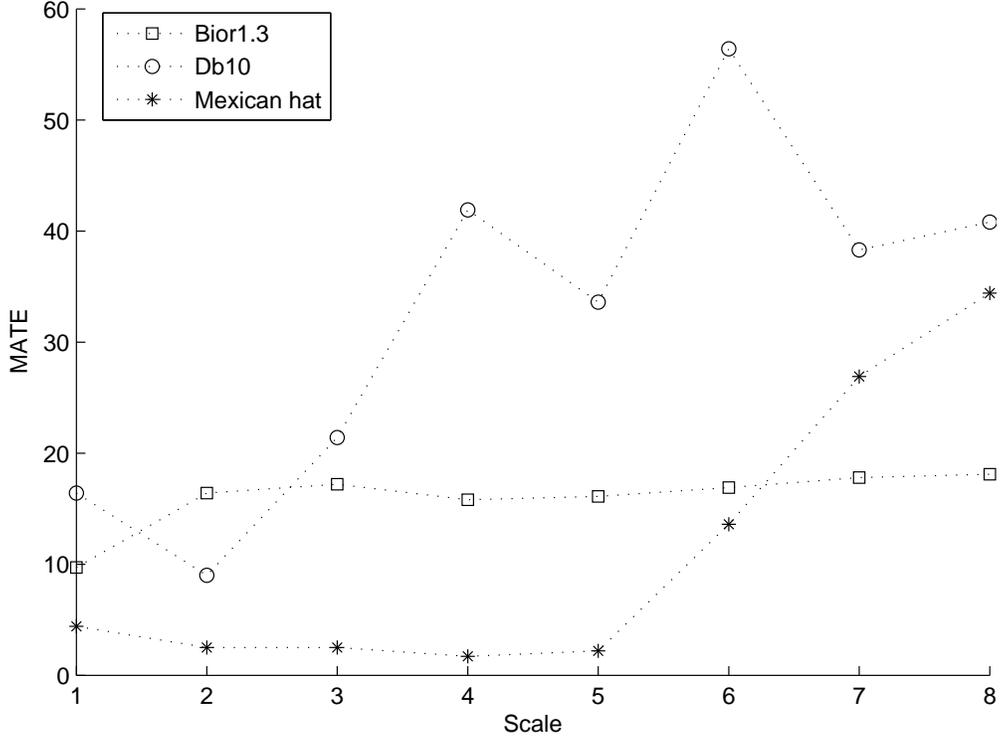}
\caption{Comparisons of the MATE value as a function of scale from three wavelet functions. Results from the Bior1.3, Db10, and Mexican hat wavelet functions are shown using square, circle, and asterisk markers, respectively.}
\label{CIN02}
\end{figure}
Fig. \ref{CIN02} shows the results of MATE as a function of the scale from 1 to 8 applied on the ECG signal record 207 from time 12.6 s to 22.6 s. Results from the Bior1.3, Db10, and Mexican hat wavelet functions are shown using square, circle, and asterisk markers, respectively. The MATE values from all wavelet functions tend to increase when the scales increase from 1 to 8. The MATE values from the Db10 wavelet function are greater than those from other wavelet functions at same scale (Except the scale 2). The maximum MATE is 56.4 ms at the scale 6 of the Db10 wavelet function and the minimum MATE is 1.7 ms at the scale 4 of the Mexican hat wavelet function.

\begin{figure}
\centering
\includegraphics[width=0.8\linewidth]{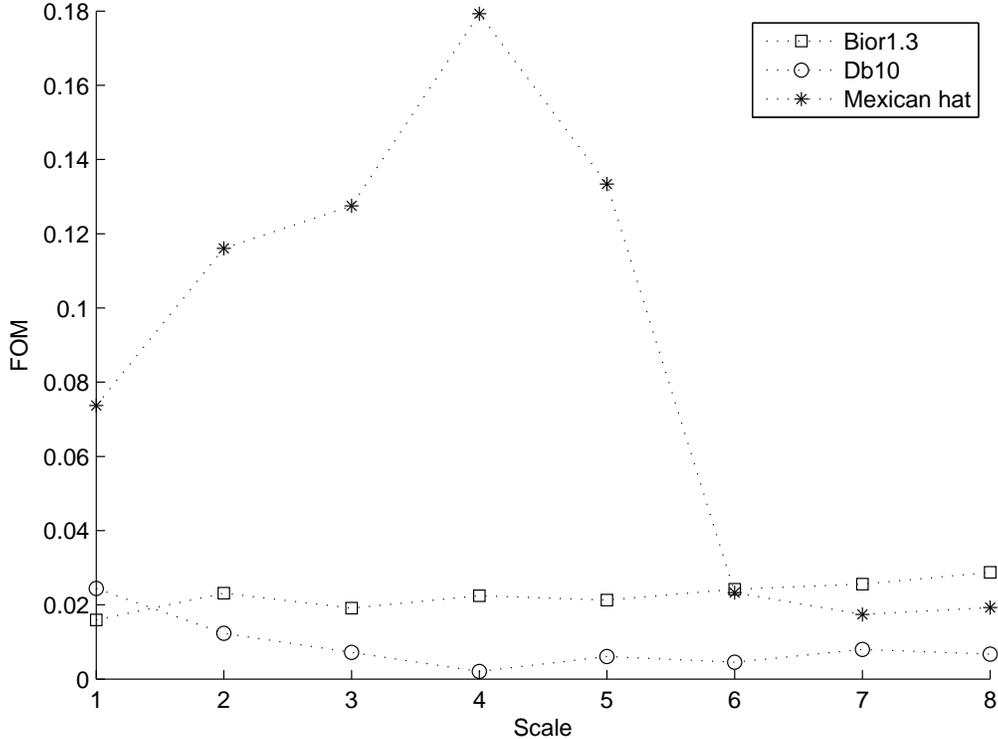}
\caption{Comparisons of the FOM value as a function of scale from three wavelet functions. Results from the Bior1.3, Db10, and Mexican hat wavelet functions are shown using square, circle, and asterisk markers, respectively.}
\label{CIN03}
\end{figure}
Fig. \ref{CIN03} shows the results of FOM as a function of the scale from 1 to 8 applied on the ECG signal record 207 from time 12.6 s to 22.6 s. Results from the Bior1.3, Db10, and Mexican hat wavelet functions are shown using square, circle, and asterisk markers, respectively. The FOM values from first five scales of the Mexican hat wavelet function are greater than those from other wavelet functions at the same scale. However, at the scale 6, 7 and 8, the FOM values from the Bior1.3 wavelet function are greater than those from other wavelet functions.    

\subsection{Signal Characteristics}
To gain more insight of the performance measurement, signal characteristics from the scale 8 of the Mexican hat wavelet function, the scale 4 of the Mexican hat wavelet function, and the scale 4 of the Db10 wavelet function are shown and discussed. While the signal characteristics from the scale 8 and 4 of the Mexican hat wavelet function are the representation for the best RMM and MATE, respectively, the signal characteristics from the scale 4 of the Db10 wavelet function are the representation of the worst FOM.  

\begin{figure}
\centering
\includegraphics[width=0.8\linewidth]{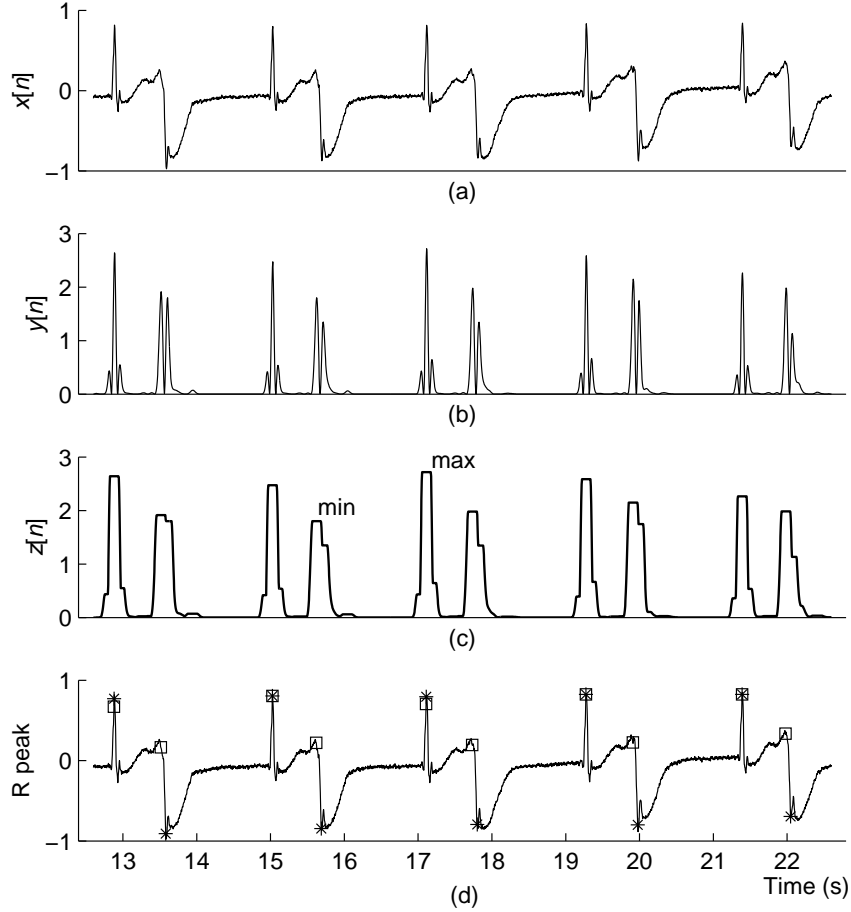}
\caption{Signal characteristics from the scale 8 of the Mexican hat wavelet function applied on the ECG signal record 207. (a) The ECG signal before noise removal $x[n]$. (b) The ECG signal after noise removal $y[n]$. (c) The envelope signal $z[n]$. (d) The ECG signal overlaid by the markers from the proposed algorithm (square) and the expert (asterisk).}
\label{CIN04}
\end{figure}
Fig. \ref{CIN04} shows signal characteristics of the proposed algorithm from the scale 8 of the Mexican hat wavelet function applied on the ECG signal record 207. Fig. \ref{CIN04}(a) shows the ECG signal before noise removal $x[n]$ from time 12.6 s to 22.6 s. The ECG signal $x[n]$ consists of 5 normal beats and 5 premature ventricular contraction beats. Fig. \ref{CIN04}(b) shows the ECG signal after noise removal $y[n]$. While the normal beat has a single peak amplitude, the premature ventricular contraction beat comprises double peak amplitudes where the left peak is greater than the right peak. Fig. \ref{CIN04}(c) shows the envelope signal $z[n]$. While the fifth beat has maximum amplitude, the forth beat has minimum amplitude. As a result, the best RMM value of 1.51 is determined. Fig. \ref{CIN04}(d) shows the ECG signal overlaid by the markers from the proposed algorithm (square) and the expert (asterisk). While all R peaks in QRS complex of normal beats are correctly detected, all R peaks in QRS complex of premature ventricular contraction beats are slightly incorrectly detected because the algorithm detects the left peak, which is greater than the right peak in all premature ventricular contraction beats. This leads to the MATE value of 34.3 ms. The corresponding FOM value in this case is 0.02.

\begin{figure}
\centering
\includegraphics[width=0.8\linewidth]{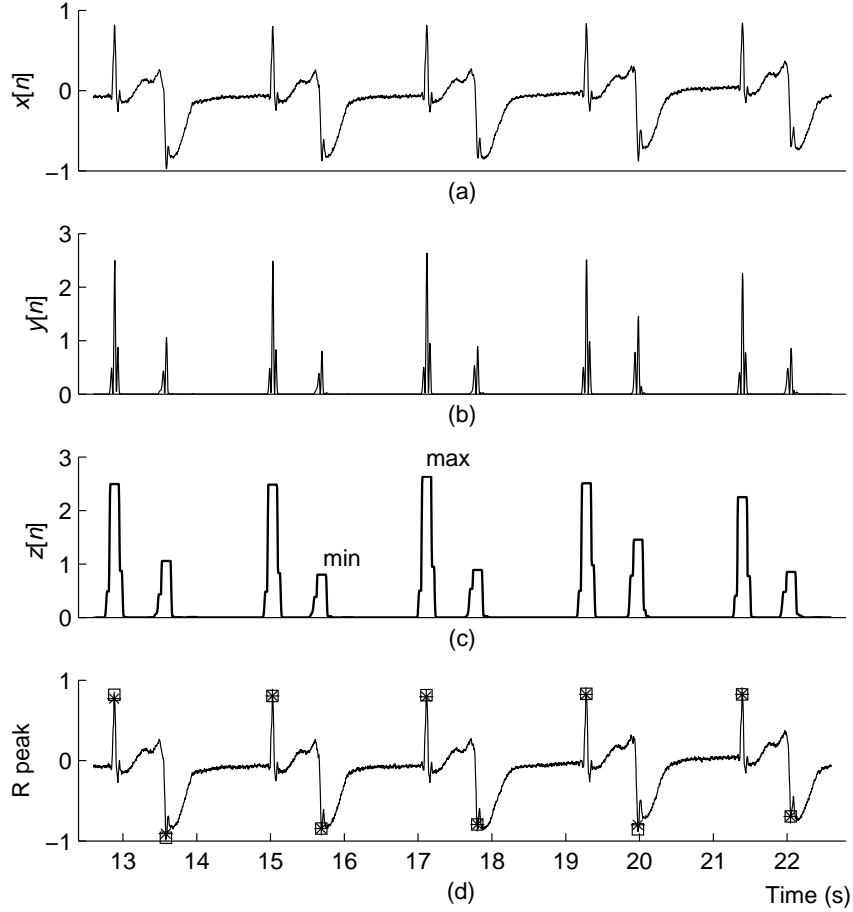}
\caption{Signal characteristics from the scale 4 of the Mexican hat wavelet function applied on the ECG signal record 207. (a) The ECG signal before noise removal $x[n]$. (b) The ECG signal after noise removal $y[n]$. (c) The envelope signal $z[n]$. (d) The ECG signal overlaid by the markers from the proposed algorithm (square) and the expert (asterisk).}
\label{CIN05}
\end{figure}
Fig. \ref{CIN05} shows signal characteristics resulting from the scale 4 of the Mexican hat wavelet function applied on the ECG signal record 207. Fig. \ref{CIN05}(b) shows the ECG signal after noise removal $y[n]$. Similar to the case in Fig. \ref{CIN04}, while the normal beat has single peak amplitude, the premature ventricular contraction beat comprises double peak amplitudes. However, the position of the larger peak amplitude in each premature ventricular contraction beat is on the right hand side. Fig. \ref{CIN05}(c) shows that the maximum beat amplitude and the minimum beat amplitude locate at the forth beat and the fifth beat, respectively. Consequently, the RMM value of 3.28 is obtained. Fig. \ref{CIN05}(d) shows the ECG signal overlaid by the markers from the proposed algorithm (square) and the expert (asterisk). All R peaks in QRS complex of both normal beats and premature ventricular contraction beats are almost correctly detected. All R peaks in QRS complex of premature ventricular contraction beats are correctly detected because the proposed algorithm detects the right peak, which is greater than the left peak in all premature ventricular contraction beats. As a result, the best value of MATE 1.7 ms is achieved. In addition, the corresponding value of FOM in this case is the best at 0.18. 

\begin{figure}
\centering
\includegraphics[width=0.8\linewidth]{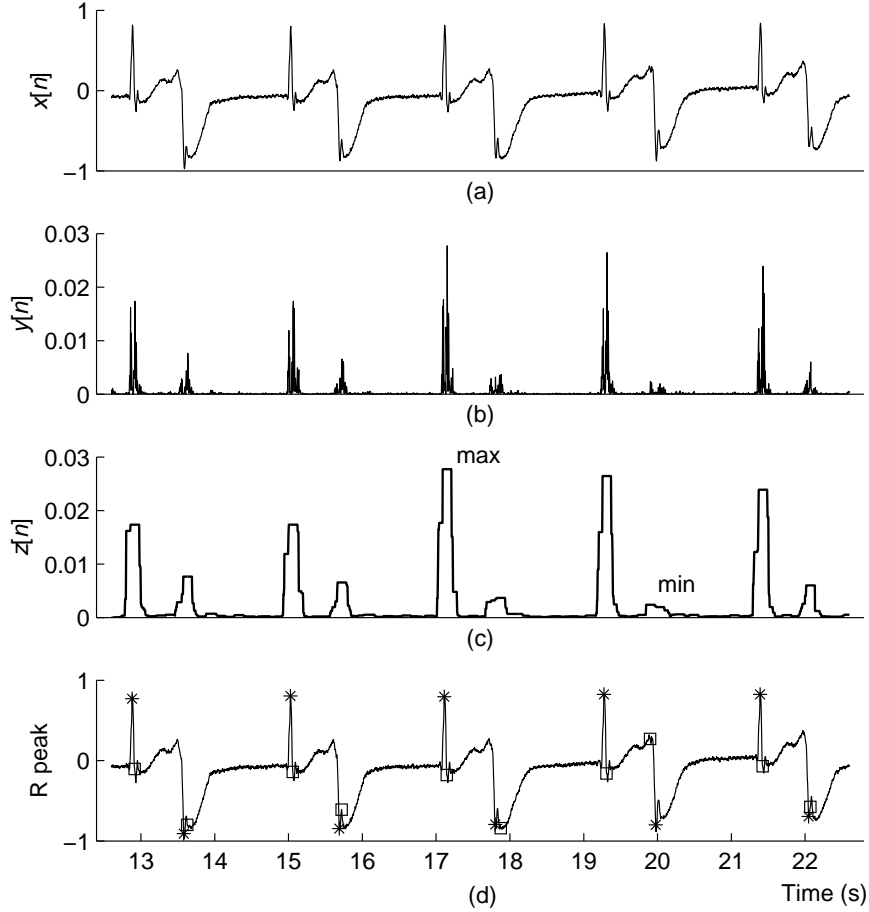}
\caption{Signal characteristics from the scale 4 of the Db10 wavelet function applied on the ECG signal record 207. (a) The ECG signal before noise removal $x[n]$. (b) The ECG signal after noise removal $y[n]$. (c) The envelope signal $z[n]$. (d) The ECG signal overlaid by the markers from the proposed algorithm (square) and the expert (asterisk).}
\label{CIN06}
\end{figure}
Fig. \ref{CIN06} shows signal characteristics resulting from the scale 4 of the Db10 wavelet function applied on the ECG signal record 207. Fig. \ref{CIN06}(b) shows the ECG signal after noise removal $y[n]$ consisting of multiple peaks, which are different from double peaks in the case of Mexican hat wavelet function. Fig. \ref{CIN06}(c) show the envelope signal $z[n]$. The position of maximum $z[n]$ and the position of minimum $z[n]$ are at the fifth beat and the eighth beat, respectively. This results in the worst RMM value of 11.69. Fig. \ref{CIN06}(d) shows the ECG signal overlaid by the markers from the proposed algorithm (square) and the expert (asterisk). All R peaks in QRS complex of both normal beats and premature ventricular contraction beats are incorrectly detected. Therefore, the high value of MATE at 41.9 ms is obtained. Moreover, the corresponding value of FOM in this case is the worst at 0.002.  

\subsection{Performance Comparisons}
To demonstrate the potential of selected wavelet functions, we apply the QRS detection algorithm to the whole ECG data of record 207. Table \ref{perEva} shows the performance comparisons of the algorithm with that from other 3 papers (\cite{14} \cite{13} \cite{8}), which use the wavelet transform and single level adaptive thresholding techniques in QRS detection algorithms. Note that each R peak from the proposed algorithm is considered as a correct detection when it locates within $\pm$50 ms from the position of the R peak given by the expert. The DER value from the scale 4 of the Mexican hat wavelet function (1.08\%) is smaller than that from the scale 8 of the Mexican hat wavelet function (1.29\%). Results show that the DER values from both wavelet functions are in the same range as those from other publications. However, the proposed algorithm uses only a single fixed threshold without any additional post-processing techniques. 

\begin{table}
\caption{Performance comparisons of the algorithm from scale 4 and 8 of the Mexican hat wavelet functions with those from other 3 publications.} \label{perEva} 
\begin{ruledtabular}
\begin{tabular}{llllllll}
Method & Total & TP  & FN  & FP  & SEN(\%) 
& PPR(\%) & DER(\%) \\
\colrule
 Scale 4  & 1860 & 1852 & 8  & 12 & 99.57 & 99.36 & 1.08 \\
 Scale 8  & 1860 & 1843 & 17 & 7  & 99.09 & 99.62 & 1.29 \\
 Choi \cite{14} & 1860 & 1848 & 12 & 10 & 99.35 & 99.46 & 1.18 \\
 Chen \cite{13} & 1863 & 1860 & 3  & 24 & 99.84 & 98.72 & 1.45 \\
 Zidelmala \cite{8}& 1872 & 1860 & 12 & 8  & 99.36 & 99.57 & 1.08 \\
\end{tabular}
\end{ruledtabular}
\label{table1}
\end{table}
     
\section{Conclusions}
We compare the capability of wavelet functions used for noise removal in QRS detection algorithm. The effects of the wavelet function on the performance in terms of QRS signal to noise ratio enhancement and detection accuracy are carefully studied and evaluated using three measurement parameters, i.e., RMM, MATE, and FOM. Three wavelet functions from previous publications are explored consisting of the Bior1.3, Db10, and Mexican hat wavelets functions. Results show that the Mexican hat wavelet function is the most appropriate for the QRS detection algorithm because it can give good results in terms of both QRS signal to noise ratio enhancement (scale 8 of the Mexican hat wavelet function) and detection accuracy (scale 4 of the Mexican hat wavelet function), which opens the opportunity for the use of a single level fixed threshold for all records of ECG data. 

To achieve a better performance, the results suggest the combination of multiple scale wavelet functions instead of using a single scale wavelet function as published in previous literature. In other words, the combination of multiple scale wavelet functions may simultaneously allow for both the enhancement QRS signal to noise ratio and the increase in detection accuracy.

\begin{acknowledgments}
This research project was funded by a grant from the Thailand Research Fund and Faculty of Engineering, Prince of Songkla University through Contract No. RSA5680043.
\end{acknowledgments}

\end{document}